%% file: main.tex
\documentclass[conference]{IEEEtran}
\usepackage{cite}
\usepackage{amsmath,amssymb,amsfonts}
\usepackage{algorithmic}
\usepackage{graphicx}
\usepackage{textcomp}
\def\BibTeX{{\rm B\kern-.05em{\sc i\kern-.025em b}\kern-.08em
    T\kern-.1667em\lower.7ex\hbox{E}\kern-.125emX}}

\usepackage{listings}
\usepackage{supertabular}
\usepackage{hyperref}
\input{macros}
\mylisting

\newlength{\xfigwd}
\begin{document}

\title{An Executable Formal Model of the VHDL in Isabelle/HOL}

\author{\IEEEauthorblockN{Wilayat Khan}
\IEEEauthorblockA{
\textit{COMSATS University Islamabad-Wah Campus}\\
Pakistan \\
wilayat@ciitwah.edu.pk}
\and
\IEEEauthorblockN{Zhe Hou}
\IEEEauthorblockA{
\textit{Griffith University}\\
Australia \\
z.hou@griffith.edu.au}
\and
\IEEEauthorblockN{David Sanan}
\IEEEauthorblockA{
\textit{Singapore Institute of Technology}\\
Singapore \\
sanan.baena@gmail.com}
\and
\IEEEauthorblockN{Jamel Nebhen}
\IEEEauthorblockA{
\textit{Prince Sattam bin Abdulaziz University}\\
Saudi Arabia \\
j.nebhen@psau.edu.sa}
\and
\IEEEauthorblockN{Yang Liu}
\IEEEauthorblockA{
\textit{Nanyang Technological University}\\
Singapore \\
yangliu@ntu.edu.sg}
\and
\IEEEauthorblockN{Alwen Tiu}
\IEEEauthorblockA{
\textit{Australian National University}\\
Australia \\
alwen.tiu@anu.edu.au}
}






\maketitle

\begin{abstract}
In the hardware design process, hardware components are usually described in a hardware description language. Most of the hardware description languages, such as Verilog and VHDL, do not have mathematical foundation and hence are not fit for formal reasoning about the design. To enable formal reasoning in one of the most commonly used description language VHDL, we define a formal model of the VHDL language in Isabelle/HOL. Our model targets the functional part of VHDL designs used in industry, specifically the design of the LEON3 processor's integer unit. We cover a wide range of features in the VHDL language that are usually not modelled in the literature and define a novel operational semantics for it. Furthermore, our model can be exported to OCaml code for execution, turning the formal model into a VHDL simulator. We have tested our simulator against simple designs used in the literature, as well as the div32 module in the LEON3 design. The Isabelle/HOL code is publicly available: \url{https://zhehou.github.io/apps/VHDLModel.zip}
\end{abstract}



\section{Introduction}
\input{intro}
\label{sec:intro}

\section{Related Work}
\input{related}
\label{sec:related}

\section{Core syntax}
\input{core_syn}
\label{sec:core_syn}

\section{Core semantics}
\input{core_sem}

\label{sec:core_sem}

\section{Semantics for Simulation}
\input{sim}
\label{sec:formsim}

\section{Components}
\input{comp}
\label{sec:comp}

\section{Complex syntax}
\input{comp_syn}

\label{sec:comp_syn}

\section{Experiment and Testing}
\input{exp}
\label{sec:exp}

\section{Conclusion}
\input{conc}
\label{sec:conc}

\bibliographystyle{IEEEtran}
\bibliography{main}

\end{document}

%% file: macros.tex
\newcommand{\mylisting}{
  \lstset{ %
	basicstyle= \footnotesize,      
	numbers=left,                   
	numberstyle=\tiny, 
	stepnumber=1,                   
	numbersep=10pt,                 
	showstringspaces=false,         
	showtabs=false,                 
	frame= l,           			
	breaklines=true,        		
	breakatwhitespace=false,    	
	escapeinside={\%*}{*)},         
	xleftmargin=.40in,				
	columns=fullflexible,  			
	mathescape,						
	framesep=5pt					
  }
}

%% file: intro.tex
VHDL is one of the most widely used hardware description languages in hardware specification, verification and documentation. However, VHDL is known to have a partially blurred semantics which is defined in plain English~\cite{vhdl2009,Reetz1995}. Formal verification, on the
other hand, is usually performed in logic. To close this gap, a formal model for VHDL is needed to verify properties of interest for hardware designs.

As a concrete motivation, this research work is a step towards building a verified execution stack ranging from CPU, micro-kernel, libraries, to applications. We are interested in verifying correctness and security properties for those components. Since the complexity of our intended goal is high, we use a multi-layer verification approach where we would formalise each layer separately and would use a refinement-based approach
to show that important properties proved at the top level
(applications) are preserved to the bottom level (CPU). We choose to formalise the XtratuM~\cite{xtratum} micro-kernel that runs on top of a multi-core LEON3 processor~\cite{leon3}, which is designed in VHDL.
A formal model for VHDL is thus vital for the low-level verification in a verified execution stack. 

We build our formal model in theorem prover Isabelle/HOL in three layers. In the bottom layer, the syntax in our model is influenced by the model of Umamageswaran et al.~\cite{Umamageswaran99}, except that 1) we focus on a synthesisable subset of VHDL while they model a timed language for simulation; and that 2) we model sub-program calls, which are not treated by Umamageswaran et al. We identify key concepts in the VHDL design of LEON3 and give a ``core syntax'' from which more complicated language constructs can be obtained. This is similar to the dynamic model in~\cite{Umamageswaran99}, however, they modeled VHDL in denotational semantics, whereas we give a novel operational semantics for VHDL, called ``core semantics'', which essentially converts VHDL statements to Isabelle/HOL functions. This idea is similar to the ACL2 model for VHDL~\cite{Georgelin02}, but we model many features that are missing in their work, such as sensitivity list for processes, loops, etc. To support hierarchical designs and compositional verification, the next layer extends the core with the syntax and semantics for \emph{components} ~\cite{rodrigues2000}. The top layer further extends the model with the necessary VHDL features used in the LEON3 design and translates the more complex syntax to the core syntax. As everything is modelled in Isabelle/HOL, we do not rely on external tools to perform heavy
translation tasks, only a simple, mostly syntactical conversion from VHDL to Isabelle/HOL is required, which is much easier to handle.

This work is a part of a research project called Securify, which aims to verify an execution stack ranging from CPU, micro-kernel, libraries to applications. The project adopts a multi-layer verification approach where we formalise each layer separately and use a refinement-based approach to show that properties proved at the top level are preserved at the lower levels. This work closely connects with the other components of the project such as the formal modelling and verification of verilog~\cite{KhanSHY19} and the SPARCv8 instruction set architecture for the LEON3 processor~\cite{HouSTLH16,HouSTLHD21}, a verification framework for concurrent C-like programs~\cite{SananZHZTL17}, and automated reasoning techniques for separation logic~\cite{HouCGT18,HouSTL17,HouT16}. For easy integration, these related sub-projects partly determine our software choices such as Isabelle/HOL and hardware choices such as LEON3 and VHDL.

The rest of the paper is organized as follows. In the next section, the core syntax of our language is defined. The core semantics of the language is defined in the Section \ref{sec:core_sem}. To simulate designs, a formal semantics is defined for it in Section \ref{sec:formsim}. Different components of the architecture are formalized in Section \ref{sec:comp}. The complex syntax of the language is given in the Section \ref{sec:comp_syn}. A detailed experimental analysis is carried out in Section \ref{sec:exp}. A literature survey is included in the related work Section \ref{sec:related} and the paper is concluded in the Section \ref{sec:conc}.
 


%% file: related.tex
There are a number of papers on formalising hardware description languages in theorem prover. Braibant et al. \cite{braibant2013formal} defined  a simplified version of the language Bluespec in theorem prover Coq. Their simplified version of Bluespec, called Fe-Si, is deeply embedded language. In a recent effort, we defined a domain-specific language dubbed as VeriFormal \cite{khan2019embedding}. The language VeriFormal is a formal version of Verilog which is deeply embedded in Isabelle/HOL. It is available with a translator that translates Verilog designs into the syntax of VeriFormal. As the syntax of this language has been defined in a functional style, an automatically extracted version is executable, hence servers as the simulator with formal foundation.

Similarly, there are formalized versions of the  VHDL, however, some are less relevant
as they focus on timed VHDL models (e.g.,~\cite{bara2010}) or they focus on the theory behind formal semantics rather than the
mechanisation of it, whereas we are mainly interested in formalising
the functional part of the LEON3 design in a theorem prover. We focus
on a synthesisable subset of VHDL which does not involve statements
such as wait and delayed assignment.
Eisenbiegler et al. gave a formal model for a synchronous VHDL subset
called $\mathcal{A}\mathcal{B}\mathcal{C}$-VHDL~\cite{eisenbiegler95},
which divides VHDL statements in three types: $\mathcal{A}$
statements, including null, variable assignment, and signal
assignment, never reach a wait statement during execution;
$\mathcal{B}$ statements sometimes reach a wait; while $\mathcal{C}$
statements, namely wait statements, always reach a wait. The authors
modelled VHDL statements as functions that describe the transition
from one clock tick to the next, they also implemented a translation
from their model to HOL. Goldschlag surveyed and formalised a few
important VHDL concepts, including signal assignments for both timed
and untimed models, delta delays, resolution functions, components,
and a few extensions~\cite{goldschlag94}.  Breuer et
al.~\cite{Breuer97} proposed a refinement calculus for VHDL,
effectively reducing the verification of VHDL to a problem in temporal
logic. Their model handles signal assignments, wait, null, if, while,
and process statements at its core, and they gave a denotational
semantics for their language. For some mechanised examples, van Tassel
embedded the simulation cycle of a VHDL subset called Femto-VHDL in
HOL~\cite{tassel1993}. The Femto-VHDL subset contains simplified
conditional statements and signal assignments (with delay) for the
sequential part, and process statements for the concurrent part. Bawa
and Encrenaz gave a VHDL translation to Petri
nets~\cite{Bawa96}. Their model, although does not support features
such as subprogram calls and components, does include most of features
surveyed in related work and has a rather strong tool support. Ralf
Reetz's deep embedding of VHDL into HOL~\cite{Reetz1995} covers a
significant subset of VHDL and includes the elaboration and execution
processes. Kloos and Breuer's book gives an excellent review of
related work in that era~\cite{Kloos95}.

Two other VHDL models are worth mentioning: Umamageswaran et al.'s
book~\cite{Umamageswaran99} documented their VHDL model in PVS. Their
syntax covers a rich subset of VHDL, for which they gave a
denotational semantics, as they mainly concern timed models. Their
model is divided in two layers: a static model which covers a
complicated syntax; and a dynamic model which is a much simpler
subset. They gave a reduction algebra to covert a static model to a
dynamic model. Their model is capable of proving some interesting
properties, such as the equivalence of two VHDL designs. Another
deeply developed work is the ACL2 model for
VHDL~\cite{Georgelin02,Borrione2001,rodrigues2000}. The ACL2 model focuses on a
synthesisable subset of VHDL, which is very close to our line of
work. This model can handle some rather involved examples, such as
modules to compute factorial and power. Furthermore, the authors also
extended their work to cover components in VHDL. This is an important
step, as it enables compositional verification.

The above work laid a solid foundation for the research in this area.
However, this cannot be used directly in the projects like verifiable execution stack for two major
reasons: First, most of the related papers were published in the
1990s and their detailed reports and source code could not be retrieved. Some authors confirmed with us that their
source code was lost. Second, most of the related work uses a rather
``abstract'' syntax. Moreover, many models assume that the VHDL code
is elaborated. This is nice when demonstrating the technique, but real
industrial designs often contain many features that are not covered by
those models, such as assignments with a range specification (rarely
supported, except by~\cite{Nicoli94,Borrione2001} etc.) and
``others'', vector member access, records, types in the
std\_logic\_1164 library, cases, among many others. One can argue that
these features can be translated to some of the previous formal
models, but it would have required to verify
the translation or the elaboration process, which may not be
straightforward. Therefore, while the related work focus on simplified
models and elegant theories, we go on the opposite direction and model
VHDL with complicated features used in industry designs.

%% file: core_syn.tex
In this section, we identify a core subset of VHDL as the base of our model. This subset can be extended with many features that are widely-used in LEON3 designs. For space reasons, the reminder of this paper only introduces our model at a high level and all the definitions are not expanded and explained.

Our core model captures the basic VHDL types (boolean, bit, char, integer, positive, natural, real, time, positive, natural, string, bitstr, boolstr) and
operations over these types (logical, relational, shift, and arithmetic operations). Our language also supports other widely-used VHDL types such as \texttt{signed}, \texttt{unsigned}, \texttt{std\_logic}, \texttt{std\_ulogic}, \texttt{std\_logic\_vector} and \texttt{std\_ulogic\_vector}. These types are modelled by built-in Isabelle/HOL types and the operations are modelled as Isabelle/HOL functions. This is similar to the treatment in the ACL2 model~\cite{Borrione2001}. We define \texttt{expression} in Figure~\ref{fig:syntax-exp}, where \texttt{e} is a shorthand for \texttt{expression}.

\begin{figure} [h]
  \label{fig:syntax-exp}
\begin{tabular}{l@{\hskip 0px}l}
  \centering
  \texttt{datatype expression = } & \\
  \texttt{ uexp uop e} & unary expressions\\
  \texttt{|bexpl e lop e} & logical expressions\\
  \texttt{|bexpr e rop e} & relational expressions\\
  \texttt{|bexps e sop e} & shifting expressions\\
  \texttt{|bexpa e aop e} & arithmetic expressions\\
  \texttt{|exp\_sig signal} & signals\\
  \texttt{|exp\_prt port} & ports\\
  \texttt{|exp\_var variable} & variables\\
  \texttt{|exp\_con const} & constants\\
  \texttt{|exp\_nth e e} & get the nth member of \\
                         & a vector\\
  \texttt{|exp\_sl e e e} & get a subvector of a vector\\
  \texttt{|exp\_tl e} & convert an expression to \\
  					  & a vector\\
  \texttt{|exp\_trl e} & convert an expression to \\
  					   & a reversed vector\\
  \texttt{|exp\_r rhsl} & records
\end{tabular}
\caption{Syntax of expression.}
\end{figure}

In \texttt{exp\_nth}, the first expression must be of a vector type and the second expression must have the type natural. In \texttt{exp\_sl}, the first expression must be a vector, the last two must be naturals respectively indicating the index of the first element in the subvector and the length of the subvector. We introduce
the last two types of expressions because VHDL overloads the vector concatenation operator and the append operator. In Isabelle/HOL, when appending a list of type \texttt{'a list} to an element of type \texttt{'a}, we explicitly convert the element to a singleton list,
then concatenate it with another list. The two types of vectors are distinguished: a list is used for big endian vectors (corresponds to \texttt{to}) ; a reversely stored list is used for little endian vectors (corresponds to \texttt{downto}). The last type of expression
is for record types of signals, ports and variables. We deal with record types as lists. For example, a signal record instance corresponds to a list of signals or nested signal records as its members. Members of a record can be accessed by checking their \texttt{name}s, which are string identifiers. A similar treatment is implemented for variable records. Record types of signals and ports are inductively defined as follows:

\vspace{10px}
\begin{tabular}{ll}
  \centering
  \texttt{datatype spl = }&\\
  \quad\quad\texttt{spl\_s signal }&\\
  \quad\texttt{| spl\_p port} &\\
  \quad\texttt{| spnl "(name $\times$ (spl list))"}
\end{tabular}
\vspace{10px}

The above definition of \texttt{expression} does not include functions. Our core model treats functions as a type of statements instead and we restrict
them to the form of ``variable := function call''. The core syntax includes the sequential statements in Figure~\ref{fig:syntax-seq}.

\begin{figure}
  \label{fig:syntax-seq}
\begin{tabular}{l@{\hskip 0px}l}
  \centering
  \texttt{datatype seq\_stmt =} & \\
  \texttt{ sst\_sa name sp\_lhs amst\_rhs} & signal assignment\\
  \texttt{|sst\_va name v\_lhs asmt\_rhs} & variable assignment\\
  \texttt{|sst\_if name condition} & if statement\\
   \texttt{"seq\_stmt list" "seq\_stmt list"} \\
  \texttt{|sst\_l  name condition} & while loop\\
   \texttt{"seq\_stmt list"} \\
  \texttt{|sst\_fn name v\_clhs} & function call\\
   \texttt{ subprogcall} \\
  \texttt{|sst\_rt name asmt\_rhs} & return statement\\
  \texttt{|sst\_pc name subprogcall} & procedure statement\\
  \texttt{|sst\_n name name condition} & next statement\\
  \texttt{|sst\_e name name condition} & exit statement\\
  \texttt{|sst\_nl} & null statement\\
\end{tabular}
\caption{Syntax of sequential statements.}
\end{figure}

Every statement has a \texttt{name}, which is an identifier of type string. The left hand side \texttt{sp\_lhs} (resp. \texttt{v\_lhs}) of an assignment may be a signal/port (resp. variable) possibly with a \texttt{discrete\_range}. The right hand side \texttt{asmt\_rhs} is either an \texttt{expression} or of the form \texttt{others => expression}. In the if statement, the \texttt{condition} is a boolean expression, the two \texttt{seq\_stmt   list}s are for the ``then'' part and the ``else'' part respectively. The \texttt{seq\_stmt list} in the while loop is the loop part. In function calls and procedure calls, \texttt{subprogcall} is defined as \texttt{subprogcall = "(name $\times$ (v\_clhs list) $\times$ type)"}, where \texttt{name} is the string identifier for the subprogram, \texttt{v\_clhs list} is the list of arguments, which can only be variables in the core model, and \texttt{type} is the
return type of function calls (not used in procedure calls). Return statements are only used in functions, they simply return the \texttt{asmt\_rhs} part, which is later assigned to the \texttt{v\_clhs} part in the function call. In next and exit statements, the first \texttt{name} is the identifier of the statement, and the second \texttt{name} is the identifier of the loop statement to be exited.

As in most previous VHDL models, our core model only considers one
type of concurrent statements: process statement. Other concurrent
statements can be translated to this one, as will be shown in
Section~\ref{sec:comp_syn}. Process statements are defined as below.

\vspace{10px}
\begin{tabular}{ll}
  \centering
  \texttt{datatype conc\_stmt = } &\\
  \quad\texttt{cst\_ps name sensitivity\_list} & \\
  \quad\texttt{"seq\_stmt list"} &\\
\end{tabular}
\vspace{10px}

\noindent Note that we support a list of signals/ports as the
\texttt{sensitivity\_list} in the process statement. Models without
this, e.g.,~\cite{Georgelin02}, are restricted to activate the process
with a single signal/port, which may not be practical.

Finally, a VHDL file corresponds to a model of type
\texttt{vhdl\_desc}, which is a tuple of \texttt{(environment $\times$
  res\_fn $\times$ conc\_stmt\_list $\times$ subprogram list)}, where
\texttt{environment} is a record containing the list of signals/ports,
variables, and types; \texttt{res\_fn} are the resolution functions;
\texttt{conc\_stmt\_list} is the list of concurrent statements in the
code; and \texttt{subprogram list} is the list of subprograms
(functions and procedures) in the design.

%% file: core_sem.tex
After the core syntax of the language is defined, the core semantics is created to interpret expressions written following the syntax of the language. A \texttt{vhdl\_state} is a record consisting of the
following fields (we omit the types below):

\vspace{10px}
\begin{tabular}{ll}
  \centering
  \texttt{state\_sp} & current value for signals/ports\\
  \texttt{state\_var} & current value for variables\\
  \texttt{state\_eff\_val} & effective values for signals/ports\\
  \texttt{state\_dr\_val} & driving values for signals/ports\\
  \texttt{next\_flag} & flag for the nearest next statement\\
  \texttt{exit\_flag} & flag for the nearest exit statement\\
\end{tabular}
\vspace{10px}

\noindent This definition is inspired by the dynamic model
in~\cite{Umamageswaran99}. We distinguish the current value, effective
value, and driving value of signals/ports. In the VHDL
LRM~\cite{vhdl2009}, the effective value is ``the value obtainable by
evaluating a reference to the signal within an expression''. Taking a
cue from~\cite{Umamageswaran99}, the effective value of a signal/port
is computed using the driving values contributed by every process
statement that drives the signal. The driving value of a signal is
defined as ``the value that the signal provides as a source of other
signals''~\cite{vhdl2009}. In~\cite{Umamageswaran99}, the driving
value of a signal/port, contributed by every process statement that
drives the signal, is computed by passing the initial value of the
signal through the list of sequential statements of the process. These
will be detailed in the semantics.

The operational semantics for if statements is straightforwardly
modeled as conditional statements in Isabelle/HOL. While loops, however,
requires some care to accommodate \texttt{next} and \texttt{exit} in
loops. The \texttt{next\_flag} and \texttt{exit\_flag} in the state
are both of type \texttt{name $\times$ bool}, the former records the
identifier of the loop, the latter is set to true when a next/exit
statement is executed. The execution of a loop, where \texttt{p} is
the current process statement, \texttt{s} is the current sequential
statement, \texttt{subps} are the subprograms in the VHDL design, and
\texttt{state} is the current state,
is modelled in Figure~\ref{fig:exec-loop-stmt}.

\begin{figure}
  \label{fig:exec-loop-stmt}
\begin{tabular}{ll}
  \texttt{"exec\_loop\_stmt p s subps state = (} &\\
  \textbf{case} \texttt{s} \textbf{of} \texttt{(sst\_l n c ssl) => (} &\\
  \quad\textbf{if} \texttt{snd (exit\_flag state) $\land$} &\\
  \quad\quad\texttt{(fst (exit\_flag state) = n)} &\\
  \quad\textbf{then} \texttt{state(|exit\_flag := (' ' ' ',False)|)} &\\
  \quad\textbf{else if} \texttt{snd (exit\_flag state)} \textbf{then} \texttt{state} &\\
  \quad\textbf{else if} \texttt{snd (next\_flag state) $\land$} &\\
  \quad\quad\texttt{(fst (next\_flag state) = n)} &\\
  \quad\textbf{then} \texttt{rec\_loop p s} &\\
  \quad\quad\texttt{(state(|next\_flag := (' ' ' ',False)|))} &\\
  \quad\textbf{else if} \texttt{snd (next\_flag state)} \textbf{then} \texttt{state} &\\
  \quad\textbf{else} \texttt{rec\_loop p s state))"} &\\
\end{tabular}
\caption{The definition of exec\_loop\_stmt.}
\end{figure}

The \textbf{if} part means that there is an exit flag active for this
loop, so we reset flag to false and change nothing else in the
state, i.e., exit this loop. The first \textbf{else if} indicates that
an exit flag is active but is not for this loop, that is, we need to
exit an outer level loop. So we exit the current loop by simply
returning \texttt{state}. The second \textbf{else if} means that a
next flag is active for the current loop, so we execute the loop again
from beginning by invoking the function \texttt{rec\_loop}, and reset
the next flag. The third \textbf{else if} says that a next flag is
active, but it is for an outer level loop. So we exit the current loop
without resetting the next flag. In the \textbf{else} case, we execute
the current loop. The function \texttt{rec\_loop} first checks the
loop condition, if the condition holds, then we sequentially execute the
statements in \texttt{s} of process \texttt{p}, and go back to \texttt{exec\_loop\_stmt}.
Note that a next/exit statement not only
sets the flags, but also ignores the remaining statements in the loop,
and calls \texttt{exec\_loop\_stmt}.

As mentioned earlier, in the core model we only support function calls
of the form ``v := function call''. This allows us to model function
calls and procedure calls in a similar way. For a function call
\texttt{sst\_fn n v spc}, where \texttt{n} is the name, \texttt{v} is
the variable on the left hand side of the assignment, and \texttt{spc}
is the function call, we first match \texttt{n} with the names of
subprograms in \texttt{vhdl\_desc}, and find the function to be
called. Since all variables are globally visible in our model, we can
pass arguments and return values via variable assignments. For
example, for a function \texttt{f(x;y)}, which is called by
\texttt{f(i,j)} with arguments \texttt{i,j}, we create assignments
\begin{tabular}{ll}
  \texttt{x := i; y := j}
\end{tabular}
and execute them before executing the function. We then execute the
body of the function and obtain an expression \texttt{e} from the
return statement. Lastly, we create an assignment \texttt{v := e},
which will be executed after the function execution is
finished. Although rarely used in the LEON3 design, it is possible to
define recursive functions, in which case function arguments
(\texttt{i,j} in the above example) will be overwritten in nested
function executions in our model. To solve this, we execute the
function body by passing a copy of the current state as a parameter,
thus we can retrieve the values of function local variables from the
original copy of the current state.

In case of procedure calls, where there are no return values, we
create a variable assignment for each \texttt{out} direction parameter
and execute these assignments after the procedure execution is
finished. For a procedure \texttt{p(x : in; y : out; z : out)} (we
omit types here), which is called by \texttt{p(i,j,k)}, we execute an
assignment \texttt{x := i} before executing the procedure, and execute
assignments \texttt{j := y; k := z} after executing the procedure.

Compared to the operational semantics for programming languages, a
major difference in an operational semantics for VHDL is in the signal
assignment statement, in which we assign the right hand side to the
driving value of the signal. The field \texttt{state\_dr\_val} in a
state has the type \texttt{sigprt => conc\_stmt => val option}, where
\texttt{sigprt} is either a signal or a port, \texttt{conc\_stmt}
corresponds to a process, which is the only type of concurrent
statement in the core model, and \texttt{val option} is either
\texttt{Some} value or \texttt{None}. That is, each driving value of a
signal is tied to a process that drives it. Due to the already rather
involved syntax, the semantics for signal assignments considers a
number of cases.
\begin{itemize}
\item Assignments for signals and ports of a record type are
  translated to assignments for each member in the record.
\item If the left hand side is a signal/port \texttt{sp} which may or
  may not be a vector, and it does not specify a range, we consider
  the following cases:
  \begin{itemize}
    \item If the right hand side is an expression \texttt{e}, we first
      evaluate the expression using a function
      \texttt{state\_val\_exp\_t}, and then assign the value as the
      driving value of the signal/port for the current process. This
      is realised as follows:
      
      \begin{tabular}{ll}
        \centering
        \texttt{state(|state\_dr\_val := } & \\
        \quad\texttt{(state\_dr\_val state)} & \\
        \quad\texttt{(sp := ((state\_dr\_val state) sp)} & \\
        \quad\texttt{(p := state\_val\_exp\_t e state))|)} & \\
      \end{tabular}

    \item If the right hand side has the form \texttt{others => e},
      then we need to make a list (using the function
      \texttt{mk\_list}) where each member is the value of \texttt{e},
      and assign this list as the driving value of the left hand side
      for the current process. In this case, we replace the last line
      of the above case with below, where \texttt{vv} is the value of
      \texttt{e}, \texttt{length vl} is the length of the vector on
      the left hand side, and \texttt{val\_list} is simply the
      constructor for vector values:

      \begin{tabular}{ll}
        \centering
        \texttt{(p := Some (val\_list }&\\
         \quad\quad\quad\texttt{(mk\_list vv (length vl)))))|)}
      \end{tabular}
      
  \end{itemize}

\item If the left hand side specifies a range, then the signal/port
  \texttt{sp} must be a vector. The sub-cases here are handled
  similarly as the above cases, but we need to make sure that only the
  elements in the range are modified, and the reminder of the vector
  stays unchanged.
  
\end{itemize}

We compute the effective value of a signal/port based on its driving values.
This is implemented in Figure~\ref{fig:effective}, where \texttt{sp} is a
signal/port and \texttt{desc} is the VHDL model.

\begin{figure}
\begin{tabular}{ll}
  \centering
  \textbf{let} \texttt{drivers = get\_drivers sp desc state} \textbf{in} &\\
  \textbf{if} \texttt{drivers = []} \texttt{then} &\\
  \quad\texttt{(state\_sp state) sp} &\\
  \textbf{else if} \texttt{List.length drivers = 1} \textbf{then} &\\
  \quad\texttt{Some (hd drivers)} &\\
  \textbf{else} &\\
  \quad\textbf{case} \texttt{(fst (snd desc)) sp} \textbf{of} \texttt{Some rf =>} &\\
  \quad\quad\texttt{Some (rf drivers)|None => None} &\\
\end{tabular}
\caption{The algorithm to obtain effective value from driving value.}
\label{fig:effective}
\end{figure}

If a signal/port has no drivers, then its value is always the initial
value, which must be its current value. If it has exactly one driver,
then the effective value is the driving value. Otherwise, we resolve
the driving values using a resolution function \texttt{rf}. Unresolved
signals/ports have the value \texttt{None}.

Variable assignments are similar to signal/port assignments, except
that variables do not have driving values and effective values, we
only record the current value of variables.

%% file: sim.tex
In a simulation cycle, we execute all active processes in a sequential
order. This order should have no effect on the outcome. A process is
active if there is a signal/port in its sensitivity list that has been
changed since the last execution, i.e., its current value differs from
its effective value. We then compute new effective values and check
active signals after this round of computation. The function
\texttt{update\_sigprt} copies the effective values of signals/ports
to their current values. After a round of execution, if a process's
sensitivity list has an active signal/port, this process is then
resumed and executed again. The cycle ends when all the
processes' sensitivity lists do not have any active signals/ports. This
is realised below, where \texttt{sps} is a list of active
signals/ports.

\vspace{10px}
\begin{tabular}{ll}
  \centering
  \textbf{function} \texttt{resume\_processes} \textbf{where} &\\ 
  \texttt{"resume\_processes desc sps state = (} &\\
  \quad\textbf{let} \texttt{state1 = } &\\
  \quad\quad\texttt{exec\_proc\_all (snd (snd desc))} &\\
  \quad\quad\quad\quad\quad\quad\quad\texttt{sps state;} &\\
  \quad\quad\texttt{state2 = comp\_eff\_val } & \\
  \quad\quad\texttt{(env\_sp (fst desc)) desc state1;} &\\
  \quad\quad\texttt{act\_sp1 = active\_sigprts desc state2;} &\\
  \quad\quad\texttt{state3 = update\_sigprt } & \\
  \quad\quad\texttt{(env\_sp (fst desc)) desc state2} \textbf{in} &\\
  \quad\textbf{if} \texttt{has\_active\_processes desc act\_sp1} \textbf{then} &\\
  \quad\quad\texttt{resume\_processes desc act\_sp1 state3} &\\
  \quad\textbf{else} \texttt{state3)"}
\end{tabular}
\vspace{10px}

\noindent Executing a simulation cycle consists of checking the active
signals/ports in each process, and executing a process if it has active
signals/ports. This is modeled as follows:

\vspace{10px}
\begin{tabular}{ll}
  \centering
  \textbf{definition} \texttt{exec\_sim\_cyc} \textbf{where} & \\
  \texttt{"exec\_sim\_cyc desc state $\equiv$} & \\
  \quad\textbf{let} \texttt{act\_sp = active\_sigprts desc state} \textbf{in} &\\
  \quad\textbf{if} \texttt{has\_active\_process desc act\_sp} \textbf{then} & \\
  \quad\quad\texttt{resume\_processes desc act\_sp state} & \\
  \quad\textbf{else} \texttt{state"} & \\
\end{tabular}
\vspace{10px}

\noindent The semantics for simulation is a straightforward recursive function:

\vspace{10px}
\begin{tabular}{ll}
  \centering
  \textbf{fun} \texttt{simulation} \textbf{where} & \\
  \texttt{"simulation 0 desc state = state"} & \\
  \texttt{|"simulation n desc state = } &\\
    \quad\texttt{simulation (n-1) desc} & \\
    \quad\texttt{(flip\_clk (exec\_sim\_cyc desc state))"} & \\
\end{tabular}
\vspace{10px}

\noindent Since most designs use a clock signal to synchronise certain
processes, we simulate the flip of a clock by the function
\texttt{flip\_clk}. 

\paragraph*{Example:} The VHDL code below demonstrates the difference between VHDL semantics and common semantics for programming languages.

\vspace{10px}
\begin{tabular}{ll}
  \centering
  \textbf{process} \texttt{(M, N)} &\\
  \textbf{begin} &\\
  \quad\texttt{M <= 1; N <= 2; X <= M + N; } &\\
  \quad\texttt{M <= 3; Y <= M + N;} &\\
  \textbf{end} &\\
\end{tabular}
\vspace{10px}

\noindent Suppose the value of \texttt{M} and \texttt{N}, before the process is
executed, are both 0. Since each signal/port only has one driving value
from each process, the second assignment of \texttt{M} overwrites the
driving value of \texttt{M} from this process. At the end of this
assignment sequence, the driving values of \texttt{M,N,X,Y} are
respectively 3, 2, 0, 0. Note that the old values (which are the
current effective values) of \texttt{M,N} are used in the assignment
of \texttt{X,Y}. The driving values only become effective after the
process suspends. Then, the effective values of \texttt{M,N} have been
changed from 0s to 3, 2 respectively, triggering the process to resume
and execute again. After the second execution finishes, the final
effective values of \texttt{M,N,X,Y} are 3, 2, 5, 5, and the process
does not have any active signals, thus this simulation cycle ends with
\texttt{X,Y} having the same value.

%% file: comp.tex
Architectures with components is widely-used in industrial designs and 
is the key to support compositional verification. Extending the
core model, a VHDL design with components, corresponds to a list of
pairs \texttt{(name $\times$ vhdl\_desc)}, where \texttt{name} is a
string identifier and \texttt{vhdl\_desc} is the model for the
component. Like in the ACL2 model~\cite{rodrigues2000}, we handle
components by giving each component a state. The state for an
architecture with components is defined inductively as below:

\vspace{10px}
\begin{tabular}{ll}
  \centering
  \texttt{datatype vhdl\_arch\_state =} &\\
  \texttt{arch\_state "(name $\times$ vhdl\_state $\times$} &\\
  \texttt{(compo\_port\_map $\times$ vhdl\_arch\_state) list)"} &\\
\end{tabular}
\vspace{10px}

\noindent Here, each component in the architecture corresponds to a
\texttt{(compo\_port\_map $\times$ vhdl\_arch\_state)} pair, the
former of which is a mapping from component ports to the outer-level
architecture ports, the latter is the state for the component. The
simulation of an architecture with components is captured by the
following functions, whose types are omitted:

\vspace{10px}
\begin{supertabular}{ll}
\textbf{function (sequential)} &\\
  \quad\quad\texttt{sim\_arch} \textbf{and} \texttt{sim\_comps} \textbf{where} &\\
  \texttt{"sim\_arch 0 descs state = state"} &\\
  \texttt{|"sim\_arch n descs state = (} &\\ 
  \textbf{let} \texttt{this\_desc = get\_desc descs state} \textbf{in} &\\
  \textbf{case} \texttt{this\_desc} \textbf{of} \texttt{Some d => (} &\\
  \quad\textbf{case} \texttt{state} \textbf{of} \texttt{arch\_state s =>} &\\
  \quad\quad\textbf{if} \texttt{(snd (snd s)) = []} \textbf{then} &\\
  \quad\quad\quad\texttt{sim\_arch (n-1) descs }&\\
  \quad\quad\quad\texttt{(arch\_state ((fst s),} &\\
  \quad\quad\quad\texttt{(simulation 1 d (fst (snd s))),[]))} &\\
  \quad\quad\textbf{else} &\\
  \quad\quad\quad\textbf{let} \texttt{new\_cl = pass\_input\_all\_comps} &\\
  \quad\quad\quad\quad\quad\quad\quad\texttt{descs state (snd (snd s));} &\\
  \quad\quad\quad\quad\texttt{sim\_cl = sim\_comps descs new\_cl;} &\\
  \quad\quad\quad\quad\texttt{output\_s = }&\\
  \quad\quad\quad\quad\quad\texttt{get\_comp\_results\_state descs} &\\
  \quad\quad\quad\quad\quad\texttt{(arch\_state ((fst s),} &\\
  \quad\quad\quad\quad\quad\texttt{(fst (snd s)),sim\_cl));} &\\
  \quad\quad\quad\quad\texttt{ns = arch\_state ((fst s),}&\\
  \quad\quad\quad\quad\quad\texttt{(simulation 1 d} &\\
  \quad\quad\quad\quad\quad\texttt{(state\_of\_arch output\_s),} &\\
  \quad\quad\quad\quad\quad\texttt{(comps\_of\_arch output\_s)));} &\\
  \quad\quad\quad\textbf{in} \texttt{sim\_arch (n-1) descs ns)} &\\
  \texttt{|None => state)"} &\\
  \texttt{|"sim\_comps descs [] = []"} &\\
  \texttt{|"sim\_comps descs (x\#xs) = }&\\
  \quad\texttt{((fst x),(sim\_arch 1} &\\
  \quad\texttt{descs (snd x)))\#(sim\_comps descs xs)"} &\\
\end{supertabular}
\vspace{10px}

\noindent If \texttt{(snd (snd s)) = []}, then this architecture does
not have component instances, thus we only have to simulate it as in
the core model for 1 cycle. Otherwise, we first pass the input from
the outer architecture to each component, obtaining a new list of
component states \texttt{new\_cl}. Each component is then simulated
for 1 cycle, yielding a new list of component states
\texttt{sim\_cl}. Next, the output of each component is passed to the
outer architecture, giving a new state \texttt{output\_s} for the
outer architecture. Finally, the outer architecture is simulated for 1
cycle, giving the next state \texttt{ns}.

%% file: comp_syn.tex
Most formal models for VHDL apply on elaborated code or some
simplified syntax, and use external software to convert more
complicated syntax to the syntax accepted by the model. However, this
route may lead to a low confidence on the correctness of the formal
method, because the external software may not be formalised and may
contain errors. To partially-overcome this, we provide a layer to
extend our model with more complicated syntax. As this layer is
formalised as a part of our formal model, we can verify the
correctness of the translation in the future. Having this layer also
improves the extensibility of our model: we only show a few treatments
here, but more can be added if one wants to adopt our model in other
situations.

In addition to the core sequential statements, we further support more
complicated \texttt{if} statements with optional \texttt{elsif} parts;
\texttt{case} statements; and \texttt{for} loops. The new sequential
statement type is called \texttt{seq\_stmt\_complex}. These new
language constructs are defined as follows:

\vspace{10px}
\begin{tabular}{ll}
  \centering
  \texttt{|ssc\_if name condition }&\\
  \quad\texttt{"seq\_stmt\_complex list"} &\\
  \quad\texttt{"elseif\_complex list" }&\\
  \quad\texttt{"seq\_stmt\_complex list"} &\\
  \texttt{|ssc\_case name expressoin }&\\
  \quad\texttt{"when\_complex list"} &\\
  \quad\texttt{"seq\_stmt\_complex list"} &\\
  \texttt{|ssc\_for name expression }&\\
  \quad\texttt{discrete\_range} &\\
  \quad\texttt{"seq\_stmt\_complex list"} &\\
\end{tabular}
\vspace{10px}

\noindent where

\vspace{10px}
\begin{tabular}{ll}
  \centering
  \texttt{elseif\_complex = ssc\_elseif condition} &\\
  \quad\texttt{"seq\_stmt\_complex list"} &\\
  \texttt{when\_complex = ssc\_when choices} &\\
  \quad\texttt{"seq\_stmt\_complex list"} &\\
\end{tabular}
\vspace{10px}

\noindent In the extended \texttt{if} syntax, the first and last
\texttt{seq\_stmt\_complex list} are the ``if'' part and ``else''
respectively, and the \texttt{elseif\_complex list} is the ``else if''
part. The \texttt{case} statement matches the \texttt{expression} with
the \texttt{choices} (which is a list of expressions) in the
\texttt{when\_complex list}, and executes the corresponding list of
sequential statements when a match is found. If no matches are found,
the ``others'' part, which is the last part of the syntax, is
executed. The \texttt{for} statement executes the list of sequential
statements repeatedly while incrementing the \texttt{expression}
within the \texttt{discrete\_range}. The translation from the above
syntax to the core syntax is straightforward and is not discussed here.

We add two types of concurrent statements which are widely-used in the
LEON3 design: concurrent signal assignments and generate
statement. They have the following forms respectively, where
\texttt{conc\_stmt\_complex} is the type of the new concurrent
statement syntax:

\vspace{10px}
\begin{tabular}{ll}
  \centering
  \texttt{|csc\_ca name sp\_clhs }&\\
  \quad\texttt{"casmt\_rhs list" asmt\_rhs} &\\
  \texttt{|csc\_gen name gen\_type }&\\
  \quad\texttt{"conc\_stmt\_complex list"} &\\
\end{tabular}
\vspace{10px}

\noindent The left hand side \texttt{sp\_clhs} of the assignment can
either be a \texttt{sp\_lhs} or a \texttt{spl}. The right hand side
\texttt{asmt\_rhs} is the same as the right hand side of sequential
assignments. In the middle part, each \texttt{casmt\_rhs} is of the
form

\vspace{10px}
\begin{tabular}{l}
  \centering
  \texttt{as\_when asmt\_rhs condition},
\end{tabular}
\vspace{10px}

\noindent which corresponds to ``\texttt{asmt\_rhs} \textbf{when}
\texttt{condition} \textbf{else}'' in the VHDL syntax. As
in~\cite{Umamageswaran99}, a concurrent signal assignment is
translated to a process with signal assignments nested in \texttt{if}
statements. Consider the following concurrent signal assignment:

\vspace{10px}
\begin{tabular}{ll}
  \centering
  \texttt{s <= x} \textbf{when} \texttt{i > 0} \textbf{else} & \\
  \quad\quad\quad\texttt{y} \textbf{when} \texttt{j = 5} \textbf{else} \texttt{z} & \\
\end{tabular}
\vspace{10px}

\noindent We translate this assignment to a process statement as
below, where we put signals \texttt{i,j} in the sensitivity list of
the process:

\vspace{10px}
\begin{tabular}{ll}
  \centering
  \texttt{thisproc:} \textbf{process} \texttt{(i,j)} \textbf{begin} & \\
  \quad\textbf{if} \texttt{(i > 0)} \textbf{then} \texttt{s <= x;} & \\
  \quad\textbf{elsif} \texttt{(j = 5)} \textbf{then} \texttt{s <= y;} & \\
  \quad\textbf{else} \texttt{s <= z;} &\\
  \quad\textbf{end if;} & \\
  \textbf{end process} \texttt{thisproc;} & \\
\end{tabular}
\vspace{10px}

We consider two types (of \texttt{gen\_type}) of generate statements:
\texttt{for} generate, and \texttt{if} generate:

\vspace{10px}
\begin{tabular}{ll}
  \centering
  \texttt{for\_gen expression discrete\_range} &\\
  \texttt{|if\_gen expression}
\end{tabular}
\vspace{10px}

\noindent Unlike the usual elaboration process, we translate a
generate statement to a list of process statements. For an \texttt{if}
generation, we evaluate the \texttt{expression} and create a list of
process statements which correspond to \texttt{conc\_stmt\_complex
  list} if the expression is evaluated to be true.
The translation of \texttt{for} generations are more tricky. For
example, if the expression is \texttt{e} and the
\texttt{discrete\_range} is \texttt{1 to 10}, then we need to create
10 process statements for each member of \texttt{conc\_stmt\_complex
  list}, and globally replace \texttt{e} with the corresponding
iteration number in the process statement. For example, consider the
generate statement below, where \texttt{p1, p2} are two process
statements:

\vspace{10px}
\begin{tabular}{ll}
  \centering
  \texttt{thisgen:} \textbf{for} \texttt{i} \textbf{in} \texttt{(0} \textbf{to} \texttt{9)} \textbf{generate begin} & \\
  \quad\texttt{p1; p2;} & \\
  \textbf{end generate} \texttt{thisgen;} & \\
\end{tabular}
\vspace{10px}

\noindent We need to generate 10 process statements based on
\texttt{p1}: \texttt{p1[0/i]}, $\cdots$, \texttt{p1[9/i]}, where
\texttt{[y/x]} means that \texttt{x} is globally replaced by
\texttt{y}. Similarly, we generate 10 process statements based on
\texttt{p2}.

We also provide abbreviations for our syntax to ease the translation
process. Following is a small portion of the div32 unit in the LEON3/GRLIB
source code~\cite{leon3sourcecode}.

\vspace{10px}
\begin{tabular}{ll}
  \centering
  \texttt{divcomb :} \textbf{process} \texttt{(r, rst, divi, addout)} & \\
  \texttt{$\cdots$} & \\
  \textbf{begin} & \\
  \quad\texttt{$\cdots$} & \\
  \quad\textbf{case} \texttt{r.state} \textbf{is} & \\
  \quad\textbf{when} \texttt{"000" =>} & \\
  \quad\quad\texttt{v.cnt := "00000";} & \\
  \quad\quad\textbf{if} \texttt{(divi.start = '1')} \textbf{then} & \\
  \quad\quad\quad\texttt{v.x(64) := divi.y(32);}&\\
  \quad\quad\quad\texttt{v.state := "001";} & \\
  \quad\quad\textbf{end if;} & \\
  \texttt{$\cdots$} & \\
  \textbf{end process;} & \\
\end{tabular}
\vspace{10px}

\noindent In our Isabelle/HOL model, this piece of code is given in Figure~\ref{fig:divcomb}.

\begin{figure}[ht]
{\small
\begin{tabular}{ll}  
  \centering
  \texttt{(''divcomb'':} &\\
  \textbf{PROCESS} \texttt{((splist\_of\_spl r)@[sp\_p rst]@} & \\
  \texttt{(splist\_of\_spl divi)@[sp\_s addout])} & \\
  \textbf{BEGIN} \texttt{[} & \\
    \texttt{$\cdots$} & \\
    \texttt{('''':} \textbf{CASE} \texttt{(exp\_of\_spl (r s.''r\_state''))} \textbf{IS} \texttt{[} & \\
      \texttt{(}\textbf{WHEN}&\\
      \texttt{[elrl [(val\_c (CHR ''0'')),} &\\
      \texttt{(val\_c (CHR ''0'')),} & \\
          \texttt{(val\_c (CHR ''0''))]] => [} & \\
        \texttt{('''': (clhs\_v (lhs\_v (var\_of\_vl} &\\
        \texttt{(v v.''v\_cnt'')))) :=} & \\
        \texttt{(rhs\_e (elrl [(val\_c (CHR ''0'')),}&\\
          \texttt{(val\_c (CHR ''0'')),} & \\
          \texttt{(val\_c (CHR ''0'')),(val\_c (CHR ''0'')),}&\\
          \texttt{(val\_c (CHR ''0''))]))),} & \\
        \texttt{('''':} \textbf{IF} \texttt{(bexpr (exp\_of\_spl }&\\
         \quad\texttt{(divi s.''divi\_start'')) [=]} &\\
         \quad\texttt{(el (CHR ''1'')))} &\\
          \quad\textbf{THEN} \texttt{[} & \\
          \quad\texttt{('''': (clhs\_v (lhs\_va (var\_of\_vl }&\\
          \quad\texttt{(v v.''v\_x'')) ((en 64)} & \\
          \quad\textbf{DOWNTO} \texttt{(en 64)))) := }&\\
          \quad\texttt{(rhs\_e (exp\_nth (exp\_of\_spl} & \\
          \quad\texttt{(divi s.''divi\_y'')) (en 32)))),} & \\
          \quad\texttt{('''': (clhs\_v (lhs\_v (var\_of\_vl }&\\
          \quad\texttt{(v v.''v\_state'')))) :=} & \\
          \quad\texttt{(rhs\_e (elrl [(val\_c (CHR ''0'')),}&\\
          \quad\texttt{(val\_c (CHR ''0'')),} & \\
            \quad\texttt{(val\_c (CHR ''1''))])))} & \\
          \texttt{] []} \textbf{ELSE} \texttt{[]} \textbf{END IF}\texttt{)]),} & \\
      \texttt{$\cdots$} & \\
      \texttt{]} \textbf{END PROCESS}\texttt{),} & \\
\end{tabular}
}
\label{fig:divcomb}
\caption{The code of divcomb in Isabelle/HOL.}
\end{figure}

It is easy to observe the resemblance between our model and the actual VHDL
code. In the above case, most of the conversion is purely syntactical, except
that we use \texttt{v.x(64 downto 64)} to access the 64th element in the vector
\texttt{v.x}, as opposed to the original code \texttt{v.x(64)}.

%% file: exp.tex
We use the Isabelle/HOL code export feature to automatically extract executable OCaml
code for our model. This enables us to run our model as a VHDL simulator for
testing purposes.

For small scale examples, we have tested the VHDL code for the
factorial function in~\cite{Georgelin02}. This design consists of two
processes: \texttt{mult} models a multiplier, and \texttt{doit}
controls the computation. The next tested design is the power function
given in~\cite{rodrigues2000}. Similar to the factorial design, the
power function design has a process for multiplication and a process
which models a finite state machine to control the computation using
the multiplier process. We have also tested a variant of the power
function design in~\cite{rodrigues2000} that contains two entities,
one for computing multiplication, the other one uses the
multiplication entity as a component and computes the power of its
inputs.

A larger tested example is the div32 unit in the LEON3/GRLIB source
code~\cite{leon3sourcecode}. This unit implements a SPARCv8 compliant
64-bit by 32-bit division, which leaves no remainder and uses the
non-restoring algorithm. The VHDL code features most of the concepts
captured in our model, including (operations on) records, concurrent
assignments, signal and variable assignments, vectors, arithmetic and
logical operations, if and case statements, process and generate
statements, etc.

The all the above tested examples, our Isabelle/HOL VHDL model successfully processes the VHDL code and generates executable code in OCaml. We have then performed extensive testing on the generated OCaml program using a large number of input parameters, including corner cases. In all the tested cases, the executable program yields correct outcome of the arithmetic functions. 

%% file: conc.tex
This paper describes a formal model of the VHDL language in
Isabelle/HOL. Our model is composed of a core of the most important
syntax and semantics, and various extensions of the core model that
handle subprogram calls, components etc. for large and modular
designs. The formalisation is coded in Isabelle/HOL, which means that
this model can be used to formally prove properties (such as
correctness) for VHDL designs. Our model is carefully crafted to
support the code export feature of Isabelle/HOL. This leads to an
executable OCaml program generated from the formal model. The program
can be seen as a VHDL simulator that strictly complies with the
syntax and semantics defined in the model. We have tested our model
through this program by running design components of the LEON3
processor and checking results.